\documentclass[11pt,a4paper]{article}
\usepackage[hyperref]{emnlp-ijcnlp-2019}
\usepackage{times}
\usepackage{latexsym}
\usepackage{balance}

\usepackage{array}
\usepackage{amssymb}
\usepackage{amsmath}
\usepackage{bm}
\usepackage{color}
\usepackage{multirow}
\usepackage{subfig}
\usepackage{arydshln}
\usepackage{amsfonts}
\usepackage{graphicx} 
\usepackage{graphics}
\usepackage{CJKutf8}
\usepackage{standalone}
\usepackage{caption}
\usepackage{url}
\usepackage{textcomp}
\usepackage{graphicx}
\usepackage{autobreak}

\definecolor{zptu}{RGB}{18, 141, 21}

\aclfinalcopy 

\title{Self-Attention with Structural Position Representations}

\author{
Xing Wang \\ Tencent AI Lab \\ {\tt brightxwang@tencent.com}   \And
Zhaopeng Tu \\ Tencent AI Lab \\ {\tt zptu@tencent.com} \AND
Longyue Wang \\ Tencent AI Lab \\ {\tt vinnylywang@tencent.com} \And
Shuming Shi \\ Tencent AI Lab \\ {\tt shumingshi@tencent.com}
}

\date{}

\begin{document}
\maketitle
\begin{abstract}
Although self-attention networks (\textsc{San}s) have advanced the state-of-the-art on various NLP tasks, one criticism of \textsc{San}s is their ability of encoding positions of input words~\cite{Shaw:2018:NAACL}. In this work, we propose to augment \textsc{San}s with {\em structural position representations} to model the latent structure of the input sentence, which is complementary to the standard sequential positional representations.
Specifically, we use dependency tree to represent the grammatical structure of a sentence, and propose two strategies to encode the positional relationships among words in the dependency tree.
Experimental results on NIST Chinese$\Rightarrow$English and WMT14 English$\Rightarrow$German  translation tasks show that the proposed approach consistently boosts performance over both the absolute and relative sequential position representations. 
\end{abstract}

\section{Introduction}
In recent years, self-attention networks ~\cite[{SANs},][]{parikh2016decomposable,lin2017structured} have achieved the state-of-the-art results on a variety of NLP tasks~\cite{Vaswani:2017:NIPS,strubell2018linguistically,Devlin:2019:NAACL}. \textsc{San}s perform the attention operation under the position-unaware ``bag-of-words'' assumption, in which positions of the input words are ignored. Therefore, absolute position~\cite{Vaswani:2017:NIPS} or relative position~\cite{Shaw:2018:NAACL} are generally used to capture the sequential order of words in the sentence. However, several researches reveal that the sequential structure may not be sufficient for NLP tasks~\cite{tai2015improved,kim2016structured,shen2018ordered}, since sentences inherently have hierarchical structures~\cite{chomsky2014aspects,bever1970cognitive}.

In response to this problem, we propose to augment \textsc{San}s with {\em structural position representations} to capture the hierarchical structure of the input sentence. The starting point for our approach is a recent finding: the latent structure of a sentence can be captured by structural depths and distances~\cite{hewitt2019structural}.
Accordingly, we propose {\em absolute structural position} to encode the depth of each word in a parsing tree, and {\em relative structural position} to encode the distance of each word pair in the tree. 

We implement our structural encoding strategies on top of \textsc{Transformer}~\cite{Vaswani:2017:NIPS} and conduct experiments on both NIST Chinese$\Rightarrow$English and WMT14 English$\Rightarrow$German translation tasks. Experimental results show that exploiting structural position encoding strategies consistently boosts performance over both the absolute and relative sequential position representations across language pairs. Linguistic analyses~\cite{P18-1198} reveal that the proposed structural position representation improves the translation performance with richer syntactic information.
Our main contributions are:
\begin{itemize}
\item Our study demonstrates the necessity and effectiveness of exploiting structural position encoding for \textsc{San}s, which benefits from modeling syntactic depth and distance under the latent structure of the sentence.

\item We propose structural position representations for \textsc{San}s to encode the latent structure of the input sentence, which are complementary to their sequential counterparts.
\end{itemize}

\begin{figure*}[t]
\centering
\includegraphics[width=\textwidth]{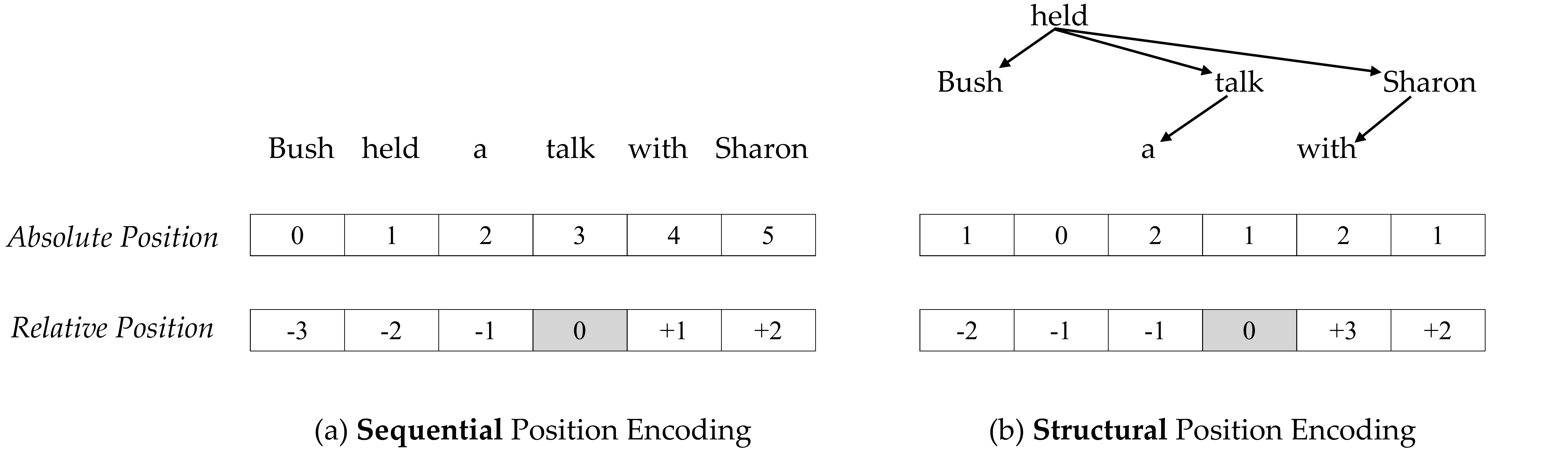}
\caption{Illustration of (a) the standard {\em sequential position encoding}~\cite{Vaswani:2017:NIPS,Shaw:2018:NAACL}, and (b) the proposed {\em structural position encoding}. The relative position in the example is for the word ``talk''.}
\label{figure:position-encoding}
\end{figure*}

\section{Background}

\paragraph{Self-Attention}
SANs produce representations by applying attention to each pair of elements from the input sequence, regardless of their distance. Given an input sequence $\mathbf{X} = \{\mathbf{x}_1,\dots,\mathbf{x}_I\} \in \mathbb{R}^{I \times d}$, the model first transforms it into queries 
$\mathbf{Q} \in \mathbb{R}^{I \times d}$, keys $\mathbf{K} \in \mathbb{R}^{I \times d}$, and values $\mathbf{V} \in \mathbb{R}^{I \times d}$:
\begin{equation}
    \mathbf{Q}, \mathbf{K}, \mathbf{V} = \mathbf{X} \mathbf{W}_Q, \mathbf{X} \mathbf{W}_K, \mathbf{X} \mathbf{W}_V \in \mathbb{R}^{I \times d},
\end{equation}
where $\{\mathbf{W}_Q, \mathbf{W}_K, \mathbf{W}_V\} \in \mathbb{R}^{d \times d}$ are trainable parameters and $d$ indicates the hidden size.
The output sequence is calculated as
\begin{eqnarray}
    \mathbf{O} &=& \textsc{Att}(\mathbf{Q}, \mathbf{K})\ \mathbf{V}  \qquad\qquad   \in \mathbb{R}^{I \times d} \label{eq:out},
\end{eqnarray}
where $\textsc{Att}(\cdot)$ is a dot-product attention model.

\paragraph{Sequential Position Encoding} To make use of the order of the sequence, information about the absolute or relative position of the elements in the sequence is injected into SAN:
\begin{itemize}
    \item {\em Absolute} Sequential PE~\cite{Vaswani:2017:NIPS} is defined as
    \begin{equation}
      \textsc{AbsPE} (abs) = f(abs / 10000^{2i/d}),  \label{eq:abs}
    \end{equation}
    where $abs$ is the absolute position in the sequence and $i$ is the dimension of position representations. $f(\cdot)$ is $\sin(\cdot)$ for the even dimension, and $\cos(\cdot)$ for the odd dimension.
    
    \newcite{Vaswani:2017:NIPS}  propose to conduct element-wise addition to combine the fixed sequential position representation with word embedding and feed the combination representation to the \textsc{Sans}. 

    \item {\em Relative} Sequential PE~\cite{Shaw:2018:NAACL} is calculated as
    \begin{equation}
      \textsc{RelPE} (rel) = \mathbf{R}[rel],  \label{eq:rel}
    \end{equation}
    where $rel$ is the relative position to the queried word, which is used to index a learnable matrix $\mathbf{R}$ that represents relative position embeddings. 
    
    \newcite{Shaw:2018:NAACL} propose relation-aware \textsc{San}s and take the relative sequential encoding as the additional key and value (Eq.\ref{eq:out}) in the attention computation.
\end{itemize}
Figure~\ref{figure:position-encoding}(a) shows an example of absolute (i.e., $abs$) and relative (i.e., $rel$) sequential positions.

\section{Approach}
\subsection{Structural Position Representations}
In this study, we choose dependency tree to represent sentence structure for its simplicity on modeling syntactic relationships among input words. Figure~\ref{figure:position-encoding} shows an example to illustrate the idea of the proposed approach.
From the perspective of relationship path between words, sequential PE measures the sequential distance between the words. As shown in Figure~\ref{figure:position-encoding} (a), for each word, absolute sequential position represents the sequential distance to the beginning of the sentence, while relative sequential position measures the relative distance to the queried word (``talk'' in the example).

The latent structure can be interpreted in various ways, from syntactic tree structures, e.g., constituency tree~\cite{collins2003head} or dependency tree~\cite{kubler2009dependency}, to semantic graph structures, e.g., abstract meaning representation graph~\cite{banarescu2013abstract}. In this work, dependency path, which is induced from the dependency tree, is adopted to provide a new perspective on modelling pairwise relationships. 

Figure~\ref{figure:position-encoding} shows the difference between the  sequential path and dependency path. The sequential distance between the two words ``held'' and ``talk'' is 2, while their structural distance is only 1 as word ``talk'' is the dependent of the head ``held''~\cite{nivre2005dependency}.

\paragraph{{\em Absolute} Structural Position}
We exploit the tree depth of the word in the dependency tree as its absolute structural position. Specifically, we treat the main verb~\cite{tapanainen1997non} of the sentence as the origin and use the distance of the dependency path from the target word to the origin as the absolute structural position
\begin{eqnarray}
      abs_{stru}(x_i) = distance_{tree}(x_i, origin),
      \label{eq:str_abs}
\end{eqnarray}
where $x_i$ is the target word, $tree$ is the given dependency structure and the origin is the main verb of the $tree$. 

In the field of NMT,  {\em BPE sub-words} and {\em end-of-sentence symbol} should be carefully handled as they do not appear in the conventional dependency tree. In this work, we assign the {\em BPE sub-words} share the absolute structural position of the original word and set the the first larger integer than the max absolute structural position in dependency tree as the absolute structural position of {\em end-of-sentence symbol}. 
\paragraph{{\em Relative} Structural Position}
For the {\em relative} structural position $rel_{stru}(x_i, x_j)$, we calculate $rel_{stru}(x_i, x_j)$ in the dependency tree following two hierarchical rules:

1. if $x_i$ and $x_j$ are at  same dependency edge, $rel_{stru}(x_i, x_j) = abs_{stru}(x_i) - abs_{stru}(x_j)$.

2. if $x_i$ and $x_j$ are at different dependency edges, $rel_{stru}(x_i, x_j) = f_{stru}(i-j) * ( abs_{stru}(x_i) + abs_{stru}(x_j))$, where
\begin{equation}
f_{stru}(x)  = \left\{
             \begin{array}{lcl}
             1 &  x > 0\\
             0 &  x = 0 \\
             -1 & x < 0 
             \end{array}
        \right.
\end {equation}

Following~\newcite{Shaw:2018:NAACL}, we use clipping distance $r$ to limit the maximum relative position.

\subsection{Integrating Structural PE into SANs}
We inherit  position encoding functions from  sequential approaches (Eq.\ref{eq:abs} and Eq.\ref{eq:rel}) to implement the structural position encoding strategies. 
Since structural position representations capture complementary position information to their sequential counterparts, we also exploit to integrate the structural position encoding into SANs with the sequential counterparts approaches~\cite{Vaswani:2017:NIPS,Shaw:2018:NAACL}. 

For the {\em absolute} position, we use the nonlinear function to fuse the sequential and structure position representations\footnote{We also use parameter-free element-wise addition method to combine two absolute position embedding and get 0.28 BLEU point improvement on development set of NIST Chinese$\Rightarrow$English over the baseline model that only uses absolute sequential encoding. }:
\begin{equation} 
\begin{split}
    asb(x_i) = &f_{abs}(\textsc{AbsPE}(abs_{seq}),\\ &\textsc{AbsPE}(abs_{stru}))
\end{split}
    \label{eqn:dec_h}
\end{equation}
where $f_{abs}$ is the nonlinear function. $\textsc{AbsPE}(abs_{seq})$ and $\textsc{AbsPE}(abs_{stru})$ are absolute sequential and structural position embedding in Eq.\ref{eq:abs} and Eq.\ref{eq:str_abs} respectively. 

For the {\em relative} position, we follow~\newcite{Shaw:2018:NAACL} to extend the self-attention computation to consider the pairwise relationships and project the relative structural position as described at Eq.(3) and Eq.(4) in~\newcite{Shaw:2018:NAACL}\footnote{Due to the space limitations we do not show  these functions. Please refer to~\newcite{Shaw:2018:NAACL} for more detail.}.  



\section{Related Work}
There has been growing interest in improving the representation power of \textsc{San}s~\cite{Dou:2018:EMNLP,Dou:2019:AAAI,Yang:2018:EMNLP,WANG:2018:COLING,Wu:2018:ACL,Yang:2019:AAAI,Yang:2019:NAACL,sukhbaatar:2019:ACL}. Among these approaches, a straightforward strategy is that augmenting the \textsc{San}s with position representations~\cite{Shaw:2018:NAACL,Ma:2019:NAACL,bello:2019:attention,Yang:2019:ICASSP}, as the position representations involves element-wise attention computation. In this work, we propose to augment \textsc{San}s  with structural position representations to  model the latent structure of the input sentence.

Our work is also related to the structure modeling for \textsc{San}s, as the proposed model utilizes the dependency tree to generate structural representations. Recently,  \newcite{Hao:2019:NAACL,Hao:2019:EMNLPb} integrate the recurrence into the \textsc{San}s and empirically demonstrate that the hybrid models achieve better performances by modeling  structure of sentences. \newcite{Hao:2019:EMNLPa} further make use of the multi-head attention to form the multi-granularity self-attention, to capture the different granularity phrases in source sentences. The difference is that we treat the position representation as a medium to transfer the structure information from the dependency tree into the  \textsc{San}s.

\begin{table}[t]
\centering
\renewcommand\arraystretch{1.1}
\scalebox{0.97}{
\begin{tabular}{c||c|c||c|c||c|c}
    \multirow{2}{*}{\bf \#} & \multicolumn{2}{c||}{\bf Sequential}	&	\multicolumn{2}{c||}{\bf Structural}    &  \multirow{2}{*}{\bf Spd.}     &   \multirow{2}{*}{\bf BLEU}\\
    \cline{2-5}
    & {\em Abs.}	&	{\em Rel.}	&	{\em Abs.}	&	{\em Rel.}  &   \\
    \hline
    1 & \multirow{3}{*}{\texttimes}	&	\multirow{3}{*}{\texttimes}	&   \texttimes  &   \texttimes  &    2.81 &  28.33\\
    2 & 	&		&   \checkmark  &   \texttimes  &   2.53 &   35.43\\
    3 &	&		&   \texttimes  &   \checkmark  &  2.65 &   34.23 \\
    \hline
    4 & \multirow{3}{*}{\checkmark}	&	\multirow{3}{*}{\texttimes}	&   \texttimes  &   \texttimes  &  3.23  &  44.31 \\
    5 &	&		&   \checkmark  &   \texttimes      &  2.65  &  44.84 \\
    6 & 	&   	&   \checkmark  &   \checkmark  &  2.52  &   45.10 \\
    \hline
    7 & \multirow{3}{*}{\checkmark}	&	\multirow{3}{*}{\checkmark}	&   \texttimes  &   \texttimes  &   3.18 &  45.02 \\
    8 & 	&		&   \checkmark  &   \texttimes  &   2.64    &  45.43 \\
    9 &  	&   	&   \checkmark  &   \checkmark  &   2.48 & 45.67 \\
\end{tabular}
}
\caption{Impact of the position encoding components on Chinese$\Rightarrow$English NIST02 development dataset using Transformer-Base model. ``Abs.'' and ``Rel.'' denote {\em absolute} and {\em relative} position encoding, respectively. ``Spd.'' denotes the decoding speed (sentences/second) on a Tesla M40, the speed of structural position encoding strategies include the step of dependency
parsing.}
\label{table-component}
\end{table}

\begin{table*}[t]
\begin{center}
\begin{tabular}{l||l l l l l|c}
      \multirow{2}{*}{\bf Model Architecture}  & \multicolumn{5}{c|}{\bf Zh$\Rightarrow$En} &  {\bf En$\Rightarrow$De}    \\
      \cline{2-7}
        & MT03 &  MT04  & MT05 & MT06 & Avg  & WMT14 \\
    \hline 
    \newcite{Hao:2019:NAACL} & - & - & - & - & -  & 28.98 \\
    \hline
     \emph{Transformer-Big}      & 45.30 & 46.49 & 45.21 & 44.87 & 45.47 & 28.58 \\
     ~~~ + {\em Structural} PE        & 45.62 & 47.12$^\uparrow$ & 45.84 & 45.64$^\Uparrow$ &  46.06 & 28.88  \\
    \hline
     ~~~ + {\em Relative} Sequential PE   & 45.45 & 47.01 & 45.65 & 45.87$^\Uparrow$ &  46.00 & 28.90 \\
     ~~~~~~~ + {\em Structural} PE        & 45.85$^\uparrow$ & 47.37$^\Uparrow$ & 46.20$^\Uparrow$ & 46.18$^\Uparrow$ & 46.40  & $\,\,$29.19$^\Uparrow$  \\
  \end{tabular}
  \caption{Evaluation of translation performance on NIST Zh$\Rightarrow$En and WMT14 En$\Rightarrow$De test sets. \newcite{Hao:2019:NAACL} is a \emph{Transformer-Big} model which adopted an additional recurrence encoder with the attentive recurrent network to model syntactic structure. ``$\uparrow/\Uparrow$'': significant over the \emph{Transformer-Big} ($p < 0.05/0.01$), tested by bootstrap resampling~\cite{koehn-2004-statistical}.} 
  \label{table:main}
  \end{center}
\end{table*}

\begin{table*}[t]
  \centering
  \scalebox{0.78}{
  \begin{tabular}{l|| c c c| c c c c| c c c c c c}
    \multirow{2}{*}{\bf Model}   & \multicolumn{3}{c|}{\bf Surface} & \multicolumn{4}{c|}{\bf Syntactic} &     \multicolumn{6}{c}{\bf Semantic}\\
    \cline{2-14}
    &\bf{SeLen} & \bf {WC} & \bf{Avg} & \bf {TrDep} & \bf {ToCo} & \bf {BShif} & \bf{Avg} & \bf {Tense} & \bf {SubN} & \bf {ObjN} & \bf {SoMo} & \bf {CoIn} & \bf{Avg} \\
    \hline
    \textsc{Base}   & 92.20 & 63.00  & 77.60 & 44.74 & 79.02 & 71.24 & 65.00 & 89.24 & 84.69 & 84.53 & 52.13 & 62.47 & 74.61\\
    \hline
    ~+ {\em Rel. Seq.} PE   
      & 89.82 &  63.17 & 76.50 & 45.09 & 78.45 & 71.40 & 64.98 & 88.74 & 87.00 & 85.53  & 51.68 & 62.21 & 75.03 \\
    \hline
    ~~~+ {\em Stru.} PE
       & 89.54 & 62.90 & 76.22 & 46.12 & 79.12 & 72.36 & 65.87 & 89.30 & 85.47 & 84.94 & 52.90 & 62.99 & 75.12\\
  \end{tabular}
  } 
  \caption{Performance on linguistic probing tasks. The probing tasks were conducted by evaluating linguistics embedded in the \emph{Transformer-Base} encoder outputs. ``Base'', ``+ {\em Rel. Seq.} PE'', ``+ {\em Stru.} PE'' denote \emph{Transformer-Base}, \emph{Transformer-Base} with relative sequential PE, \emph{Transformer-Base} with relative sequential PE and structural PE models respectively.} 
  \label{tab:probing}
\end{table*}

\section{Experiment}

We conduct  experiments  on  the  widely used  NIST Chinese$\Rightarrow$English and WMT14 English$\Rightarrow$German data, and report the 4-gram BLEU score~\cite{papineni2002bleu}.

\paragraph{Chinese$\Rightarrow$English} We use the training dataset consists of about  $1.25$ million sentence pairs. NIST 2002 (MT02) dataset is used as development set. NIST 2003, 2004, 2005, 2006 datasets are used as test sets. We use byte-pair encoding (BPE) toolkit to alleviate the out-of-vocabulary problem with 32K merge operations. 

\paragraph{English$\Rightarrow$German} We use the dataset consisting of about $4.5$ million sentence pairs as the training set. The newstest2013 and newstest2014 are used as the development set and the test set. We also apply BPE with 32K merge operations to obtain subword unit. 

We evaluate the proposed position encoding strategies on \textsc{Transformer}  \cite{Vaswani:2017:NIPS} and implement them on top of THUMT~\cite{zhang2017thumt}. We use the Stanford parser~\cite{klein2003accurate} to parse the sentences and obtain the structural structural absolute and relative position as described in Section 3. When using relative structural position encoding, we use clipping distance $r$ = 16. To make a fair comparison, we valid different position encoding strategies on the encoder and keep the \textsc{Transformer} decoder unchanged. 
We study the variations of the \textsc{Base} model on Chinese$\Rightarrow$English task, and evaluate the overall performance with the \textsc{Big} model on both translation tasks.

\subsection{Model Variations}
We evaluate the importance of the proposed absolute and relative structural position encoding strategies and study the variations with \emph{Transformer-Base} model on Chinese$\Rightarrow$English data. The experimental results on the development set are shown in Table~\ref{table-component}. 

\paragraph{Effect of Position Encoding} We first remove the sequential encoding from the \emph{Transformer} encoder (Model \#1) and observe the translation performance degrades dramatically ($28.33-44.31 = -15.98$), which demonstrates the necessity of the position encoding strategies. 

\paragraph{Effect of Structural Position Encoding} Then we valid our proposed structural position encoding strategies over the position-unaware model (Models \#2-3). We find that absolute and relative structural position encoding strategies improve the translation performance by 7.10 BLEU points and 5.90 BLEU points respectively, which shows that the introducing of the proposed absolute and relative structural positions improves the translation performance in terms of BLEU score.

\paragraph{Combination of Sequential and Structural Position Encoding Strategies} We integrate the absolute and relative structural position encoding strategies into the \emph{Base} model equipped with absolute sequential position encoding (Models \#4-6). We observe that the proposed two approaches are able to achieve improvements over the \emph{Base} model with decoding speed marginally decreases.

Finally, we valid the proposed structural position encoding over the \emph{Base} model equipped with absolute and relative sequential  position encoding (Models \#7-9). We find that sequential relative encoding obtains 0.71 BLEU points improvement (Model \#7 vs. Model \#4) and structural position encoding achieves a further improvement in performance by 0.65 BLEU points (Model \#9 vs. Model \#7), demonstrating the effectiveness of the proposed structural position encoding strategies.

\subsection{Main Results}
We valid the proposed structural encoding strategies over \emph{Transformer-Big} model in Chinese$\Rightarrow$English and English$\Rightarrow$German data, and list the results in Table~\ref{table:main}. 

For Chinese$\Rightarrow$English,  Structural position encoding (+ {\em Structural} PE) outperforms the \emph{Transformer-Big} by 0.59 BLEU points on average over four NIST test sets. Sequential relative encoding approach (+{\em Relative} Sequential PE) outperforms the \emph{Transformer-Big} by 0.53 BLEU points, and structural position encoding (+ {\em Structural} PE) achieves further improvement up to +0.40 BLEU points and outperforms the \emph{Transformer-Big} by 0.93 BLEU points.  For English$\Rightarrow$German, similar phenomenon is observed, which reveals that the proposed structural position encoding strategy can consistently boost translation performance over both the absolute and relative sequential position representations.

\subsection{Linguistic Probing Evaluation}
We conduct probing tasks\footnote{\url{https://github.com/facebookresearch/SentEval/tree/master/data/probing}}~\cite{P18-1198} to evaluate structure knowledge embedded in the encoder output in the variations of the \emph{Base} model that are trained on En$\Rightarrow$De translation task. 

We follow~\newcite{Wang:2019:ACL} to set model configurations. The experimental results on probing tasks are shown in Table~\ref{tab:probing}, and the BLEU scores of ``Base'', ``+ {\em Rel. Seq.} PE'', ``+ {\em Stru.} PE'' are 27.31, 27.99 and 28.30.   From the table, we can see 1) adding the relative sequential positional embedding achieves improvement over the baseline on semantic tasks (75.03 vs. 74.61). This may indicate the model benefits more from  semantic modeling; 2) with the structural positional embedding, the model obtains improvement on  syntactic tasks (65.87 v.s. 64.98), which indicates that  the representations preserve more syntactic knowledge.

\section{Conclusion}
In this paper, we have presented a novel  structural position encoding strategy to augment \textsc{San}s  by considering the latent structure of the input sentence.  We extract structural absolute and relative positions from the dependency tree and integrate them into \textsc{San}s.  
Experimental results on Chinese$\Rightarrow$English and English$\Rightarrow$German  translation tasks have demonstrated that the proposed approach consistently improve translation performance over both the absolute and relative sequential position representations. 

Future directions include inferring the structure representations from the AMR~\cite{song-etal-2019-semantic} or the external SMT knowledge~\cite{AAAI1714451}. 
Furthermore, the structural position encoding can be also applied to the decoder with RNN Grammars~\cite{dyer2016recurrent,eriguchi2017learning}, which we leave for future work.

\bibliography{emnlp-ijcnlp-2019}
\bibliographystyle{acl_natbib}
\end{document}